%% file: emnlp2022.tex
\pdfoutput=1

\documentclass[11pt]{article}

\usepackage[]{EMNLP2022}

\usepackage{times}
\usepackage{latexsym}
\usepackage{array}
\newcolumntype{P}[1]{>{\centering\arraybackslash}p{#1}}
\newcolumntype{M}[1]{>{\centering\arraybackslash}m{#1}}

\usepackage{fdsymbol}

\usepackage{graphicx}
\usepackage{svg}
\usepackage{comment}
\usepackage{booktabs}
\usepackage{array}
\usepackage{caption}
\usepackage{subcaption}
\usepackage{tablefootnote}
\usepackage{multirow}
\usepackage{multicol}
\usepackage{enumitem}
\usepackage{xcolor, color, soul}
\usepackage[multiple]{footmisc}

\definecolor{lightblue}{RGB}{219,226,238}
\definecolor{darkred}{RGB}{105,17,10}
\definecolor{lightyellow}{RGB}{251,242,214}
\definecolor{darkyellow}{RGB}{82,58,34}
\definecolor{lightgrey}{RGB}{230,230,230}
\definecolor{darkgrey}{RGB}{57,57,57}

\usepackage[T1]{fontenc}

\usepackage[utf8]{inputenc}

\usepackage{microtype}

\usepackage{inconsolata}

\input{commands}

\title{\synkb{}: Semantic Search for Synthetic Procedures}

\author{Fan Bai$^{\clubsuit}$ Alan Ritter$^{\clubsuit}$  Peter Madrid$^{\spadesuit}$ Dayne Freitag$^{\diamondsuit}$ John Niekrasz$^{\diamondsuit}$ \\
  $\clubsuit$ {School of Interactive Computing, Georgia Institute of Technology} \\
  $\spadesuit$ {Biosciences Division, SRI International} \\
  $\diamondsuit$ \text{Artificial Intelligence Center, SRI International} \\
 \texttt{\{fan.bai, alan.ritter\}@cc.gatech.edu} \\
 \texttt{\{peter.madrid, daynefreitag, john.niekrasz\}@sri.com} \\
 {{\bf Demo URL}: \url{https://tinyurl.com/synkb}} \\
 \\
 }

\begin{document}
\maketitle
\begin{abstract}
\input{sections/0_abstract}
\end{abstract}

\section{Introduction}
\label{sec:intro}
\input{sections/1_intro}

\section{\procsearch{}}
\label{sec:search}
\input{sections/2_search}

\section{Empirical Comparison}
\label{sec:compare}

\input{sections/3_compare}

\section{Related Work}
\label{sec:related}
\input{sections/4_related}

\section{Conclusion}
\label{sec:conclu}
\input{sections/5_conclu}

\section*{Ethical Considerations and Broader Impacts}
\label{sec:ethics}
\input{sections/7_broader_impacts}

\section*{Acknowledgements}
\label{sec:acknow}
\input{sections/6_acknow}

\bibliography{anthology,custom}
\bibliographystyle{acl_natbib}



\end{document}

%% file: commands.tex
\newcommand\procsearch{{\sc SynKB}}
\newcommand\synkb{{\sc SynKB}}

\newcommand\reaxys{Reaxys}
\newcommand\scifinder{SciFinder}

\newcommand\uspto{USPTO-Lowe}

\newcommand\chemsyn{{\textsc{ChemSyn}}}

\newcommand\chemu{{\textsc{ChEMU}}}

\newcommand\procbert{ProcBERT}

\newcolumntype{L}[1]{>{\raggedright\let\newline\\\arraybackslash\hspace{0pt}}m{#1}}
\newcolumntype{C}[1]{>{\centering\let\newline\\\arraybackslash\hspace{0pt}}m{#1}}
\newcolumntype{R}[1]{>{\raggedleft\let\newline\\\arraybackslash\hspace{0pt}}m{#1}}

\newcommand{\tabf}[1]{{\fontsize{8}{9.5}\selectfont {#1}}}

%% file: sections/0_abstract.tex
In this paper we present \synkb{},\footnote{Introduction video: \url{https://screencast-o-matic.com/watch/c3jVQsVZwOV}} an open-source, automatically extracted knowledge base of chemical synthesis protocols.
Similar to proprietary chemistry databases such as Reaxsys, \synkb{} allows chemists to retrieve structured knowledge about synthetic procedures.  By taking advantage of recent advances in natural language processing for procedural texts, \synkb{} supports more flexible queries about reaction conditions, and thus has the potential to help chemists search the literature for conditions used in relevant reactions as they design new synthetic routes.
Using customized Transformer models to automatically extract information from 6 million synthesis procedures described in U.S. and EU patents, we show that for many queries, \synkb{} has higher recall than Reaxsys, while maintaining high precision.  
We plan to make \synkb{} available as an open-source tool; in contrast, proprietary chemistry databases require costly subscriptions.\footnote{Code: \url{https://github.com/bflashcp3f/SynKB}.}

%% file: sections/1_intro.tex
Commercial chemistry databases, such as Reaxys\footnote{ \url{https://www.elsevier.com/solutions/reaxys}} are invaluable tools for chemists, who issue structured SQL-like queries to retrieve precise information about chemical reactions described in the literature.  Large, high-quality datasets are also crucial for synthetic route planning \citep{klucznik2018efficient}, automation \citep{coley2019robotic,collins2020fully}, and machine learning approaches to retrosynthesis \citep{coley2019graph}. In addition to proprietary, manually curated databases such as Reaxys, recent work has begun to use automatically extracted data from reactions described in patents \citep{tetko2020state}, however existing databases are limited to basic reaction information, and do not include important details such as concentrations or order of additions \citep{coley2019robotic}.
The lack of high-quality data has been identified as a key challenge in developing recommendation models for reaction conditions \citep{struble2020current}.


In this paper, we present \synkb{}, a working system that demonstrates the application of modern NLP methods to extract large quantities of structured information about chemical synthesis procedures from text. \synkb{} has a number of advantages with respect to existing chemistry databases such as Reaxys: (1) We show that by automatically extracting information from millions of synthesis procedures described in U.S. and European patents using state-of-the-art NLP methods, we can achieve significantly higher recall than existing chemistry databases while maintaining high precision.  In \S \ref{sec:compare}, we demonstrate \synkb{}'s coverage is complementary to Reaxys; see Figure \ref{fig:unique_answer} for details.
(2) \synkb{}'s novel graph search supports better coverage of reaction conditions than existing chemistry databases; this includes concentrations, reaction times, order of the addition of reagents, catalysts, etc.
(3) We will make \synkb{} available as open-source software on publication, in contrast, most existing chemistry databases are proprietary, with the notable exception of \citet{uspto_lowe}, which we compare to in \S \ref{sec:compare}.

We have built an online demo, which can be viewed at the following URL: \url{https://tinyurl.com/synkb}.  We will also release the source code and patent-based extractions used to build \synkb{} on publication.  

%% file: sections/2_search.tex
\begin{figure*}
  \includegraphics[width=\textwidth]{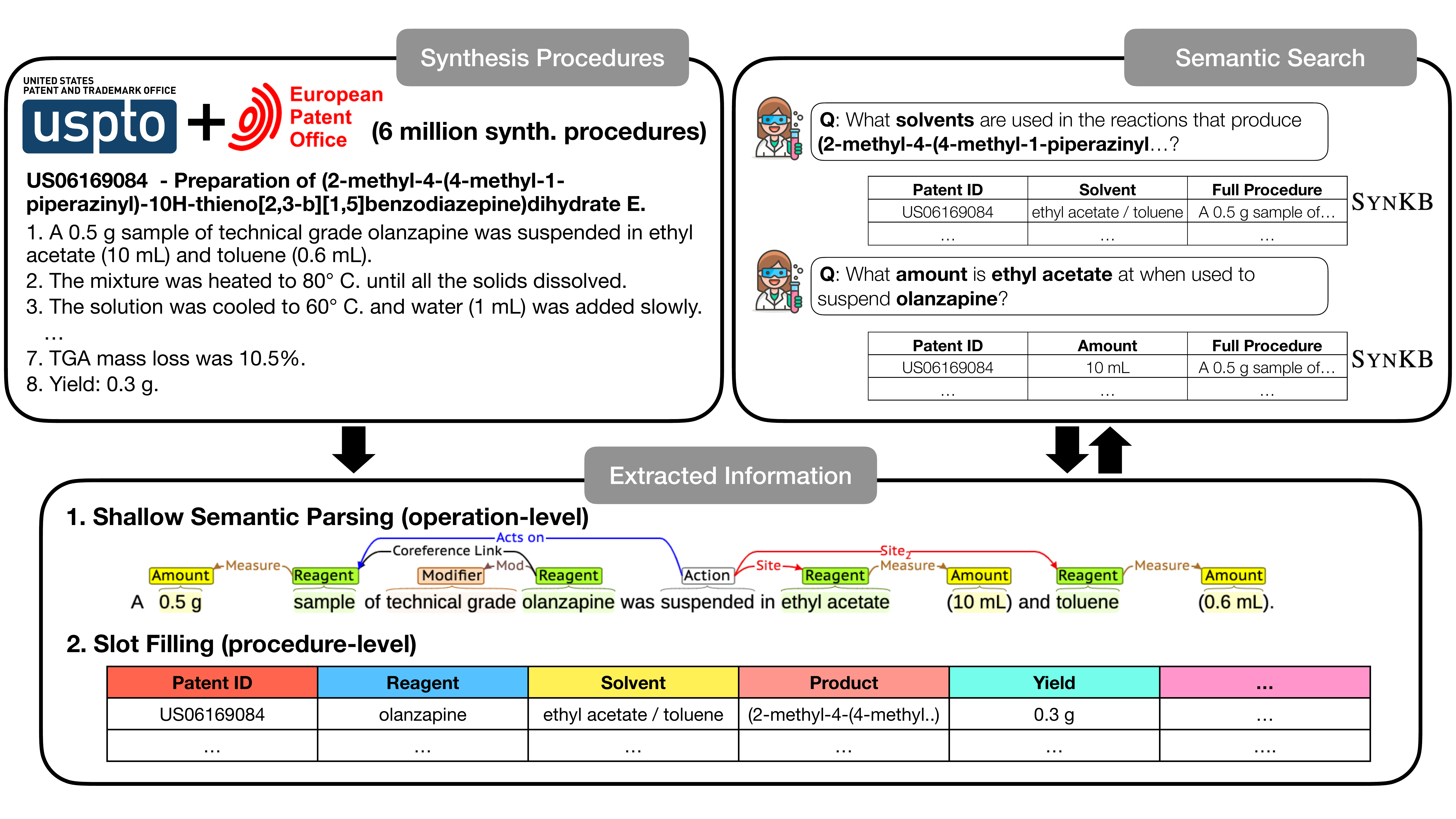}
  \caption{Overview of our semantic search system \synkb{}, which searches over 6 million chemical synthesis procedures collected from patents. Users can enter structured queries to retrieve procedures concerning procedure-level or operation-level information.}
  \label{fig:overview}
\end{figure*}

\synkb{} is an open-source system that allows chemists to perform structured queries over large corpora of synthesis procedures. 
In this section, we present each component of \synkb{}, as illustrated in Figure \ref{fig:overview}.
Our corpus collection is first presented in \S \ref{sec:data_collect}. 
Section \ref{sec:data_process} describes how a corpus of six million procedures is annotated with sentence-level action graphs, in addition to protocol-level slots relevant to chemical reactions, including starting materials, solvents, reaction products, yields, etc.  
After automatically annotating and indexing, we experiment with the semantic search capabilities enabled by \synkb{} in \S \ref{sec:sys_feature}.


\subsection{Corpus Collection}
\label{sec:data_collect}

We extract structured representations of synthetic protocols from a corpus of chemical patents \cite{bai-etal-2021-pre}, which includes over six million chemical synthesis procedures extracted from around 300k U.S. and European patents (written in English).
The U.S. portion of this corpus comes from an open-source corpus of chemical synthesis procedures \citep{uspto_lowe}, which covers 2.4 million synthetic procedures extracted from U.S. patents (USPTO\footnote{ \url{https://www.uspto.gov/learning-and-resources/bulk-data-products}}, 1976-2016).
For the European portion, we apply the  \citet{uspto_lowe} reaction identification pipeline to European patents.
Specifically, we download patents from EPO\footnote{ \url{https://www.epo.org/searching-for-patents/data/bulk-data-sets.html}} (1978-2020) as XML files and select patents containing the IPC (International Patent Classification) code ‘C07’ for processing as they are in the category of organic chemistry. 
Next, the synthesis procedure identifier developed by \citet{Lowe2012ExtractionOC}, a trained Naive Bayes classifier, is applied to the {\em Description} section of all selected patents.
As a result, we obtain another 3.7 million procedures from European patents.

\subsection{Extracting Reaction Details from Synthetic Procedures}
\label{sec:data_process}


To facilitate semantic search, we automatically annotate the corpus of 6 million synthetic procedures described above with semantic action graphs \cite{kulkarni2018annotated} in addition to chemical reaction slots \cite{Nguyen2020ChEMUNE} using Transformer models that are pre-trained on a large corpus of scientific procedures \cite{bai-etal-2021-pre}.

\paragraph{Shallow Semantic Parsing.} We first perform sentence-level annotation, where each step in the procedure is annotated with a semantic graph \citep{tamari-etal-2021-process}.  Nodes in the graph are experimental operations and their typed arguments, whereas labeled edges specify relations between the nodes (see the example shallow semantic parse in Figure \ref{fig:overview}). 
Here we use the \chemsyn{} framework \citep{bai-etal-2021-pre}, which covers 24 types of nodes (such as \textit{Action}, \textit{Reagent}, \textit{Amount}, \textit{Equipment}, etc.) and 17 edge types (e.g. \textit{Acts-on} and \textit{Measure}).
With these annotated semantic graphs, users can search for operation-level information, for example, the amount of \texttt{DMF} when used as a solvent to dissolve \texttt{HATU} (this will be further discussed in \S \ref{sec:compare}).
Following \citet{tamari-etal-2021-process}, we split semantic graph annotation into two sub-tasks, Mention Identification (MI) for node prediction and Argument Role Labeling (ARL) for edge prediction.
We use the same fine-tuning architectures as in \citet{tamari-etal-2021-process}.
Models are fine-tuned on the \chemsyn{} corpus, which consists of 992 chemical synthesis procedures extracted from patents, and the resulting performance (averages across five random seeds) is shown in Table \ref{tab:procbert_results}. We select model checkpoints via the Dev set performance out of five random seeds, and use the selected checkpoint for inference on our 6 million synthetic procedures. 

\paragraph{Slot Filling.} In the second task, we annotate procedures from a protocol perspective, i.e., identifying key entities playing certain roles in a protocol, which can be queried in a slot-based search.
We use the \chemu{} training corpus proposed in \citet{Nguyen2020ChEMUNE}.
This dataset includes 10 pre-defined slot types
concerning chemical compounds and related entities in chemical synthesis processes such as \textit{Starting Material}, \textit{Solvent}, and \textit{Product}. Similar to the Mention Identification task, we treat Slot Filling as a sequence tagging problem. However, the input in Slot Filling is the entire protocol, rather than a single sentence, as in mention identification. We fine-tune models on the \chemu{} dataset (see Table \ref{tab:procbert_results} for results), and then run inference on the chemical patent corpus using the learned model.

\paragraph{ProcBERT.} 
We use \procbert{} \citep{bai-etal-2021-pre}, a BERT-based model that is pre-trained on in-domain data (scientific protocols), as the backbone for all of our models, and develop task-specific fine-tuning architectures on top of it. The comparison between \procbert{} and other pre-trained models is presented in Table \ref{tab:procbert_results}.  Because \procbert{} is pre-trained using in-domain data, we find that it outperforms both BERT$_\text{large}$ \citep{devlin-etal-2019-bert} and SciBERT \citep{Beltagy2019SciBERT} on all three tasks.

\begin{table}[h!]
\small
\begin{center}
\scalebox{0.72}{
\begin{tabular}{lcccc}
\toprule
\multirow{2}{*}{\textbf{Annotation Task}} & \multirow{2}{*}{\textbf{Dataset}} & \multicolumn{3}{c}{\textbf{Pre-trained Model}} \\ 
 & & \textsc{BERT}\textsubscript{large} & SciBERT & \procbert{} \\
\midrule
Mention Identification & \multirow{2}{*}{\chemsyn{}} & 95.26\textsubscript{0.1} & 95.82\textsubscript{0.2} & \textbf{95.97}\textsubscript{0.2} \\ 
Argument Role Labeling  &                             & 92.87\textsubscript{0.5} & 93.27\textsubscript{0.2} & \textbf{93.57}\textsubscript{0.2} \\
 \midrule
Slot Filling                    & \chemu{}      & 95.10\textsubscript{0.2} & 95.63\textsubscript{0.1} & \textbf{96.19}\textsubscript{0.1} \\
\bottomrule
\end{tabular}
}
\end{center}
\caption{\label{tab:procbert_results} Test set F\textsubscript{1} scores of fine-tuned models for the three annotation tasks. 
These numbers, averages across five random seeds with standard deviations as subscripts, are taken from our previous work \citet{bai-etal-2021-pre}. Models using \procbert{} for contextual embeddings perform the best on all three tasks and are used for automatic annotations on six million synthesis procedures to construct \synkb{}.
}
\end{table}

\begin{table}[h!]
\small
\begin{center}
\scalebox{0.78}{
\begin{tabular}{lcccc}
\toprule

 & \textbf{\synkb{} (ours)} & \textbf{\uspto{}} & \textbf{\reaxys{}} \\
  \midrule
License & Open source & Open source & Subscription \\ 
\# Procedures (mill.) & 6 & 2.4 & 57 \\ 
\# Entity Types & 24 & 8 & 10 \\
\# Relation Types & 17 & - & - \\
Annotation & Automatic & Automatic & Manual \\
\bottomrule
\end{tabular}
}
\end{center}
\caption{\label{tab:database_comp} Comparison between our \synkb{} and two performant databases. Our \synkb{} provides more fine-grained annotations (more entity types and unique relation annotations) than the other two systems and covers more procedures than \uspto{}, a database built using the largest open-source synthesis procedure corpus \citep{uspto_lowe}.
}
\end{table}


\subsection{Semantic Search}
\label{sec:sys_feature}
\synkb{} offers search modalities specific to each of these two forms of annotation, i.e., semantic action graphs and chemical reaction slots, along with features designed to support practical use.
The first type of query supported by \synkb{} is \textbf{semantic graph search}, which allows users to search for synthesis procedures based on the semantic parse of the constituent operations. We adapt the graph query formalism proposed originally for syntactic dependencies in \citet{valenzuela-escarcega-etal-2020-odinson}.\footnote{We refer readers to the \href{https://gh.lum.ai/odinson/queries.html}{tutorial} of Odinson query language for more details of this graph query formalism.} 
Formally, the input query $G = (V, E)$ is a labeled directed graph. Each node $v_i \in V$ is specified as a set of constraints on matching entities (a single or multi-token span).
For example, users can specify the node as \texttt{DMF} or \texttt{[word=DMF]}, which triggers an exact match on entity mentions containing the word ``DMF''. 
They can also constrain the entity type of the node using the expression \texttt{[entity=Type]}.\footnote{We store entity labels with the BIO tagging scheme, so users can match a single token entity with the expression \texttt{[entity=B-Type]} and a multi-token entity with the expression \texttt{[entity=B-Type][entity=I-Type]*}.} Moreover, nodes can be named \texttt{captures} when surrounded with \texttt{(?<name>...)}, e.g., the query \texttt{(?<solvent> DMF)} captures \texttt{DMF} as the \texttt{solvent}. As for the edge $e = (v_i, v_j, l) \in E$, we need to specify the direction and the semantic relation. Considering the query \texttt{(?<solvent> DMF) >measure (?<amount> 1 ml)}, it represents a semantic graph containing two entity nodes captured as \texttt{solvent} and \texttt{amount}, and an edge signaling the \texttt{measure} relation and its direction (from \texttt{solvent} to \texttt{amount}).
In addition, \synkb{} supports \textbf{slot-based search}, which presents a structured search interface, with entries corresponding to \chemu{} slots.  A keyword entered into any entry restricts the retrieved set to procedures where the extracted slot contains the indicated keyword.  Like the graph search, this returns a set of tuples with elements named with matching slots and containing the matching entity strings.  The special token \texttt{``?''} can be used to match \emph{any} slot value. 




As for the implementation, the semantic graph search module is powered by Odinson \citep{valenzuela-escarcega-etal-2020-odinson}, an open-source Lucene-based query engine. 
Odinson pre-indexes the annotated corpus by generating the inverted index for each procedure.
Given an input query, Odinson performs a two-step matching process, where it first examines the node constraints via the inverted index; if this step works well, the semantic relations will be verified in the second step. 
The two-step matching process improves the speed of Odinson, 
and thus enables interactive querying.
As for the slot-based search, it is supported by Elasticsearch\footnote{\url{https://www.elastic.co/elasticsearch/}} with the exception that, when users perform both types of search at the same time, we use the metadata search feature of Odinson for slot filters (we store slot values as metadata) to improve the system's response speed. 


%% file: sections/3_compare.tex
\begin{table*}[th!]
\small
\centering
\scalebox{0.85}{
\begin{tabular}{lL{11cm}ccc}
\toprule 
 \tabf{\textbf{System}} & \tabf{\textbf{Input Query}} & \tabf{\textbf{\# Proce.}} & \tabf{\textbf{\# Ans.}} & \tabf{\textbf{Ans. Prec.}} \\
 
\midrule
\midrule

\multicolumn{5}{c}{\tabf{\textbf{Slot-based Search}}} \\
\midrule

\multicolumn{5}{l}{\tabf{\textbf{Q1} - What are the \textbf{solvents} used for reactions containing the reagent \textbf{triphosgene}?}} \\ 

 \tabf{\reaxys{}}  & \tabf{\texttt{\{"reagent":"triphosgene"\}}} & \tabf{35} & \tabf{7} & \tabf{\textbf{100}\%}\\


 \tabf{\uspto{}} & \tabf{\multirow{2}{=}{\texttt{\{"reagent":"triphosgene", "solvent":"?"\}}}} & \tabf{3157} & \tabf{104} & \tabf{90\%} \\
 \tabf{\synkb{}} &  & \tabf{\textbf{7184}} & \tabf{\textbf{127}} & \tabf{94\%} \\

 \midrule
 
  \multicolumn{5}{l}{\tabf{\textbf{Q2} - What are the \textbf{yields} (percent) of reactions producing \textbf{(5-Methylpyrimidin-2-yl)methanol}?}} \\
 \tabf{\reaxys{}} & \tabf{\texttt{\{"product":"(5-Methylpyrimidin-2-yl)methanol"\}}} & \tabf{1} & \tabf{1} & \tabf{100\%} \\
 \tabf{\uspto{}} & \tabf{\multirow{2}{=}{\texttt{\{"product":"(5-Methylpyrimidin-2-yl)methanol", "yield (percent)":"?"\}}}} & \tabf{1} & \tabf{1} & \tabf{100\%}\\
 \tabf{\synkb{}} &  & \tabf{1} & \tabf{1} & \tabf{100\%}\\

 \midrule

 \multicolumn{5}{l}{\tabf{\textbf{Q3} - What are the \textbf{products} of reactions containing the reagent \textbf{trimethylsilyldiazomethane}?}} \\
 \tabf{\reaxys{}} & \tabf{\texttt{\{"reagent":"trimethylsilyldiazomethane"\}}} & \tabf{438} & \tabf{75} & \tabf{\textbf{100}\%}  \\
 \tabf{\uspto{}} & \tabf{\multirow{2}{=}{\texttt{\{"reagent":"trimethylsilyldiazomethane", "product":"?"\}}}} & \tabf{517} & \tabf{335} & \tabf{98\%} \\
 \tabf{\synkb{}} &  & \tabf{\textbf{1033}} & \tabf{\textbf{708}} & \tabf{96\%} \\
 
 \midrule

 

 \multicolumn{5}{l}{\tabf{\textbf{Q4} - What are the \textbf{products} of reactions containing the reagent \textbf{chlorosulfonic acid} and the solvent \textbf{chlorobenzene}?}} \\
 \tabf{\reaxys{}} & \tabf{\texttt{\{"reagent":"chlorosulfonic acid"\} AND \{"solvent":"chlorobenzene"\}}} & \tabf{\textbf{148}} & \tabf{\textbf{65}} & \tabf{100\%} \\
 \tabf{\uspto{}} & \tabf{\multirow{2}{=}{\texttt{\{"reagent":"chlorosulfonic acid", "solvent":"chlorobenzene", "product":"?"\}}}} & \tabf{6} & \tabf{2} & \tabf{100\%} \\
 \tabf{\synkb{}} &  & \tabf{9} & \tabf{4} & \tabf{100\%} \\
 
 \midrule
 
  \multicolumn{5}{l}{\tabf{\textbf{Q5} - What are the \textbf{reaction times} for reactions using reagent \textbf{CDI (carbonyldiimidazole)}?}} \\
 \tabf{\reaxys{}} & \tabf{\texttt{\{"reagent":"CDI"\} OR \texttt{\{"reagent":"carbonyldiimidazole"\}}}} & \tabf{93} & \tabf{24} & \tabf{\textbf{100}\%} \\
 \tabf{\uspto{}} & \tabf{\multirow{2}{=}{\texttt{\{"reagent": "CDI OR carbonyldiimidazole", "reaction time":"?"\}}}} & \tabf{3722} & \tabf{339} & \tabf{100\%} \\
 \tabf{\synkb{}} &  & \tabf{\textbf{6377}} & \tabf{\textbf{511}} & \tabf{94\%} \\
 
 \midrule
 
  \multicolumn{5}{l}{\tabf{\textbf{Q6} - What are the \textbf{reaction temperatures} for reactions containing reagent \textbf{trifluoromethanesulfonic acid}?}} \\
 \tabf{\reaxys{}} & \tabf{\texttt{\{"reagent":"trifluoromethanesulfonic acid"\}}} & \tabf{104} & \tabf{3} & \tabf{\textbf{100}\%} \\
  \tabf{\uspto{}} & \tabf{\multirow{2}{=}{\texttt{\{"reagent":"trifluoromethanesulfonic acid", "temperature":"?"\}}}} & \tabf{727} & \tabf{124} & \tabf{100\%}\\
 \tabf{\synkb{}} & & \tabf{\textbf{1937}} & \tabf{\textbf{243}} & \tabf{98\%}\\
 
 \midrule
 \midrule
 
\multicolumn{5}{c}{\tabf{\textbf{Semantic Graph Search}}} \\
 
\midrule


 
 \multicolumn{5}{l}{\tabf{\textbf{Q7} - What are the \textbf{reagents} used to dilute \textbf{plasma}?}} \\
 \tabf{\synkb{}} & \tabf{\texttt{plasma <acts-on diluted >using (?<reagent> [entity=B-Reagent][entity=I-Reagent]*)}} & \tabf{24} & \tabf{16} & \tabf{100\%} \\

\midrule
 
\multicolumn{5}{l}{\tabf{\textbf{Q8} - What is the \textbf{pH} of a solution after being titrated with \textbf{NaOH}?}} \\
 \tabf{\synkb{}} & \tabf{\texttt{(?<ph> [entity=B-pH][entity=I-pH]+) <setting titrated >using NaOH}} & \tabf{39} & \tabf{21} & \tabf{95\%} \\
 
\midrule
 
 \multicolumn{5}{l}{\tabf{\textbf{Q9} - What are the common \textbf{pore sizes} of \textbf{PTFE filters}?}} \\
 \tabf{\synkb{}} & \tabf{\texttt{PTFE filter >measure (?<pore\_size> [entity=B-Generic-Measure][entity=I-Generic-Measure]*)}} & \tabf{183} & \tabf{39} & \tabf{92\%} \\
 
\midrule
 
 \multicolumn{5}{l}{\tabf{\textbf{Q10} - What \textbf{molar concentration} is the reagent \textbf{HATU} at when \textbf{dissolved} in the solvent \textbf{DMF}?}} \\
 \tabf{\synkb{}} & \tabf{\texttt{HATU >measure (?<mole> [] [word=mmol|word=mol]) []\{1,10\} DMF >measure (?<volume> [] [word=ml|word=l])}} & \tabf{447} & \tabf{289} & \tabf{100\%} \\
 
 \bottomrule
\end{tabular}
}

\caption{\label{tab:results} Search queries and resulting performance on 10 chemist-proposed questions for \reaxys{}, \uspto{}, and \synkb{} (ours). 
\textbf{\# Proc.} is the number of returned procedures containing valid answers, and \textbf{\# Ans.} refers to the number of distinct answer slots or captures in these procedures.
The first six questions (Q1-Q6) are answerable for all three databases as they only require entity annotation while the last four questions (Q7-Q10) can only be answered by our \synkb{} using our unique semantic action graph annotation.
\synkb{} consistently shows better recall than two compared databases while being highly accurate.
}
\end{table*}


In \S \ref{sec:search}, we described the design and implementation of \synkb{} including the underlying models, data preparation, and semantic search features. 
To demonstrate the utility of \synkb{} for assisting chemists to search the literature for reaction details, we now evaluate its search features on ten example questions (Q1-Q10 in Table \ref{tab:results}), which were collected from synthetic chemists working on the design of new synthesis protocols. 
In \S\ref{sec:slot_eval}, we evaluate the slot-based search module of \synkb{} and compare it with two existing databases which provides similar search features.
In \S\ref{sec:seman_eval}, we demonstrate how to use our novel semantic graph search module to answer operation-specific questions and evaluate its retrieved answers and procedures.

\subsection{Slot-based Search Evaluation}
\label{sec:slot_eval}
We benchmark the slot-based search module of \synkb{} against \reaxys{}, one of the leading proprietary chemistry databases, and \uspto{}, an automatically extracted database built using a large open-source synthesis procedure corpus \citep{uspto_lowe}.
Below, we first introduce these two databases briefly, and then evaluate the results of all three systems on the chemist-proposed questions.

\subsubsection{Chemistry Databases}
The first database we compare with is \textbf{\reaxys{}}, a web-based commercial chemistry database, which contains comprehensive chemistry data, including chemical properties, compound structures, etc.
What particularly interests us in \reaxys{} is that it contains expert-curated reaction procedures collected from extensive published literature such as chemistry-related patents and periodicals.\footnote{\url{https://www.elsevier.com/solutions/reaxys/features-and-capabilities/content}} Also, key experimental entities in those reaction procedures, like participating reagents and reaction temperature, are specified.
Thus, similar to our slot-based search, \reaxys{} allows users to search for reaction procedure information by applying text filters. Users can use its \textit{Query Builder} module to specify multiple chemical reaction-specific filters, and then \reaxys{} returns all matched reaction procedures along with identified entities in those procedures, which are available for download.
Apart from \reaxys{}, we also build a database using \textbf{\uspto{}} \citep{uspto_lowe}, the largest available open-source chemical synthesis procedure corpus 
as introduced in \S \ref{sec:data_collect}, for comparison.
Similar to our \synkb{}, this corpus includes automatic annotations of experimental entities on 2.4 million contained reaction procedures.\footnote{\url{https://www.nextmovesoftware.com/leadmine.html}}
However, our \synkb{} provides more fine-grained and comprehensive entity annotations (see Table \ref{tab:database_comp} for the statistics of three experimented databases), and also annotates the relations between extracted entities, which constitute semantic graphs (\S\ref{sec:data_process}) enabling operation-specific semantic graph search.
As for the implementation, we load \uspto{}'s entity annotation into Elasticsearch, so this customized database can be used in the same way as the slot-based search module of our \synkb{}.


\subsubsection{Comparison with Examples}
We now compare three systems on six questions that were proposed by chemists (Q1-Q6) as these questions only require annotations on experimental entities and thus can be answered in all three systems.
For example, Q1 (``What solvents are used in reactions involving triphosgene?'') can be answered by the \synkb{} query \texttt{\small\{"reagent":"triphosgene", "solvent":"?"\}}, as \textit{reagent} and \textit{solvent} are query-able ChEMU slots.
Similarly, for \reaxys{}, experimental entities are specified for corresponding text filters.

We evaluate the output of each system from two perspectives: 1) recall, which is measured by the number of returned procedures containing valid answers and the number of distinct answer slots or captures in these procedures; and 2) precision, the proportion of correct answers among all predicted answers.
In cases where the number of answers exceeds 50, we sample 50 answers from the full set to estimate precision.

\begin{figure}[!t]
    \centering
    \includegraphics[width=0.35\textwidth]{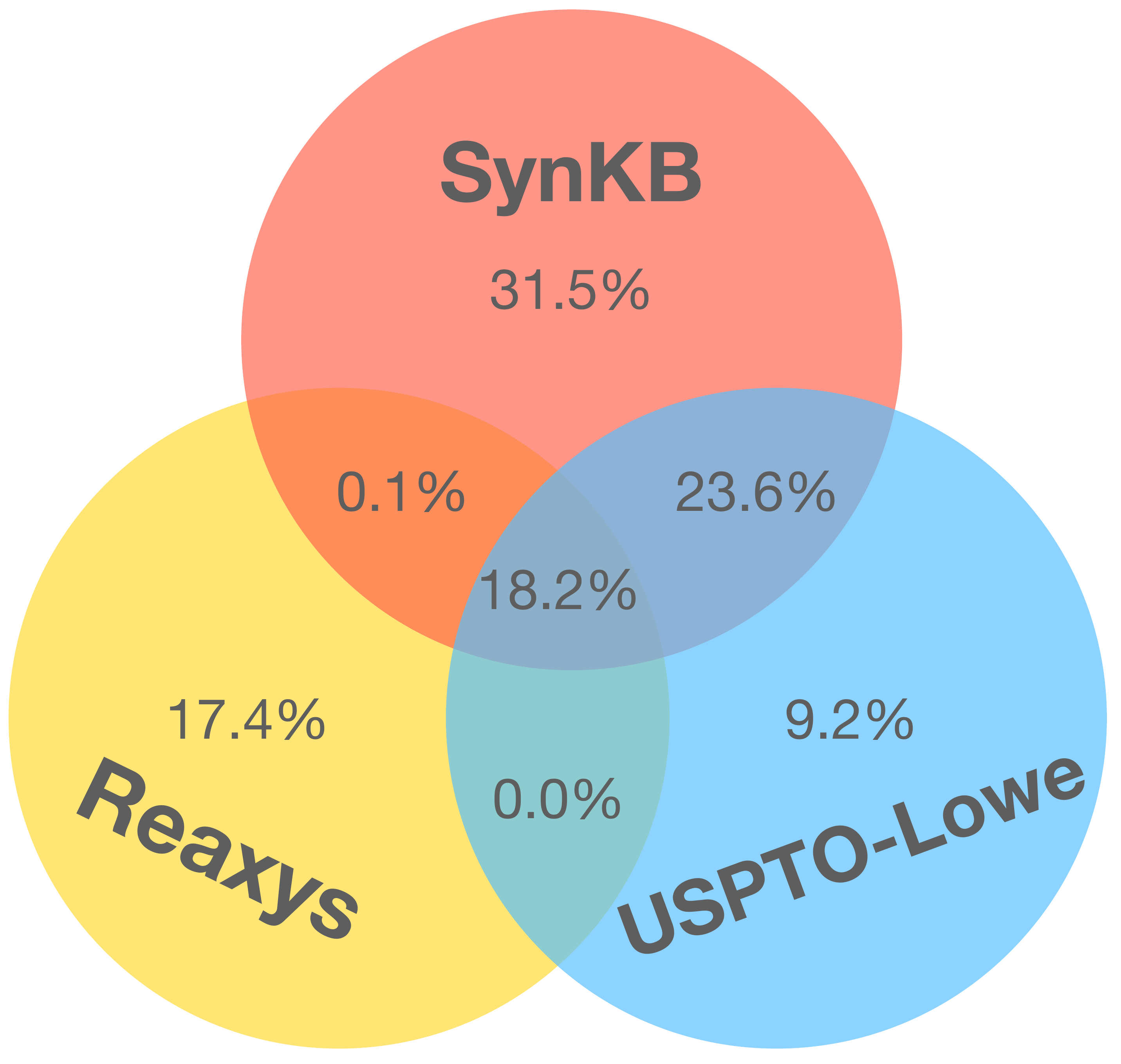}
    \caption{Venn diagram on the answer distribution of six slot-based search questions (macro-average) for all three databases. We can see that both our \synkb{} and \reaxys{} cover high percentage of unique answers, suggesting that users should use them together if possible. }
    \label{fig:unique_answer}
\end{figure}

The search queries and performance on each question for the three systems are shown in Table \ref{tab:results}.
We can see that, \synkb{} consistently retrieves a larger number of relevant procedures and answers than \reaxys{} (5 out of 6 questions) while maintaining high precision.
\uspto{}, which uses a rule-based annotation model, shows competitive performance on precision but trails behind our \synkb{} in terms of recall for all 6 questions.
This comparison clearly shows the strength of our system: by leveraging state-of-the-art NLP for chemical synthesis procedures \cite{bai-etal-2021-pre}, we can provide chemists with abundant information, which is non-proprietary and delivered with high precision.
Furthermore, we plot the Venn diagram (Figure \ref{fig:unique_answer}) over the retrieved answers, which shows the percentage of unique and shared answers for each system out of all retrieved answers (we do macro-average across six questions.)
Interestingly, only 18.1\% of retrieved answers are shared among all three systems, and both our \synkb{} and \reaxys{} contain a large number of unique answers, which take 31.5\% and 17.4\% of retrieved answers respectively. 
This shift in answer distribution suggests that our open-source \synkb{} can be a good complement to proprietary chemistry databases like \reaxys{}, and it is better for users to use both of them if possible instead of choosing one over the other.

\subsection{Semantic Graph Search Evaluation}
\label{sec:seman_eval}
We evaluate our novel semantic graph search on four operation-specific questions (Q7-Q10). Unlike the six questions introduced above, these questions place constraints on the relations between mentioned entities, and thus are not answerable for \reaxys{} and \uspto{} (due to the lack of relation annotation).
For instance, to answer Q7 ``What are the reagents used to dilute \texttt{plasma}?'', a system needs to first locate the particular operation in a procedure where \texttt{plasma} is diluted, and then identify the reagent, which facilitates this dilution operation.
This whole process can be realized in our semantic graph search module.
Concretely, the graph-based query we use for Q7 is:
\texttt{\small ``plasma <acts-on diluted >using (?<reagent> [entity=B-Reagent][entity=I-Reagent]*)''}, 
which matches procedures containing ``\texttt{plasma}'' and ``\texttt{diluted}'' connected in the same semantic graph and returns used reagents in the form of named captures. 
We evaluate the performance of the semantic graph search module by manually inspecting predicted answers (randomly sampling 50 answers for Q10), and show results in Table \ref{tab:results}. Similar to the findings in the slot-based search evaluation, \synkb{} shows good coverage while maintaining high precision.

%% file: sections/4_related.tex
\citet{Lowe2012ExtractionOC} was the first to develop a complete information extraction pipeline for chemical synthesis procedures, using a mostly rule-based approach. 
Subsequently, there have been several efforts to extract information from experimental procedures by either developing more performant extraction models \citep{vaucher2020automated, guo2021correction} or designing extraction frameworks for other types of scientific literature, 
like wet-lab protocols \citep{kulkarni2018annotated} and material science publications \citep{mysore2019materials, kuniyoshi2020annotating, olivetti2020data, o2021ms}.
In this paper, we use the state-of-the-art NLP models for chemical synthesis procedures \citep{bai-etal-2021-pre} to build the largest open-source knowledge base that searches synthetic procedure details.
Our system is complementary to many proprietary chemistry databases, such as \reaxys{}, \scifinder{}\footnote{\url{https://scifinder.cas.org}}, and Pistachio\footnote{\url{https://www.nextmovesoftware.com/pistachio.html}}, in terms of contained information and search modalities.

Recent work has also developed slot-based classifiers to extract structured representations of events (from social media), supporting structured queries \cite{zong2020extracting}.  In contrast, we present a semantic search system, which is customized for chemical synthesis procedures with specialized search features.
In addition, recent work has  explored {\em extractive search} systems \cite{ravfogel2021neural} that allow experts to specify syntactic patterns, including syntactic structures of the input and capture slots.  The graph-based queries in our \procsearch{} enable a similar capability in the domain of synthetic procedures, however \procsearch{}'s queries are defined over semantic graphs that encode actions in synthetic protocols and associated semantic arguments.


%% file: sections/5_conclu.tex
In this paper, we present \synkb{}, a system for large-scale extraction and querying of chemical synthesis procedures. 
\synkb{} provides efficient searches against semantic action graphs and chemical reaction slots derived from 6 million synthesis procedures contained in chemical patents. 
A quantitative comparison with \reaxys{}, one of the leading commercial databases of reaction information, demonstrates the competence and versatility of our freely accessible system.

%% file: sections/7_broader_impacts.tex
Proprietary chemistry databases, such as \reaxys{} require costly subscriptions, limiting scientific inquiry for those who do not have the means to access this valuable source of information.  In this paper, we presented an open-source semantic search system, \synkb{}, which demonstrates state-of-the-art NLP methods can enable automatically extracted databases of synthetic procedure operational details that are competitive with \reaxys{} in terms of recall.  We will make our code and data freely available.

The data contained in \synkb{} is based on automatic extraction from both European and U.S. patents that are in the public domain.  Our use complies with the terms of service of the U.S. Patent and Trademark Office and the European Patent Office.

%% file: sections/6_acknow.tex
We thank Wei Xu for helpful discussion and comments on a previous draft.
This material is based upon work supported by the Defense Advanced Research Projects Agency (DARPA) under Contract No. HR001119C0108.
The views, opinions, and/or findings expressed are those of the author(s) and should not be interpreted as representing the official views or policies of the Department of Defense or the U.S. Government.

%% file: emnlp2022.bbl
\begin{thebibliography}{23}
\expandafter\ifx\csname natexlab\endcsname\relax\def\natexlab#1{#1}\fi

\bibitem[{Bai et~al.(2021)Bai, Ritter, and Xu}]{bai-etal-2021-pre}
Fan Bai, Alan Ritter, and Wei Xu. 2021.
\newblock \href {https://doi.org/10.18653/v1/2021.emnlp-main.409} {Pre-train or
  annotate? domain adaptation with a constrained budget}.
\newblock In \emph{Proceedings of the 2021 Conference on Empirical Methods in
  Natural Language Processing}, pages 5002--5015, Online and Punta Cana,
  Dominican Republic. Association for Computational Linguistics.

\bibitem[{Beltagy et~al.(2019)Beltagy, Lo, and Cohan}]{Beltagy2019SciBERT}
Iz~Beltagy, Kyle Lo, and Arman Cohan. 2019.
\newblock \href {https://doi.org/10.18653/v1/D19-1371} {{S}ci{BERT}: {A}
  {P}retrained {L}anguage {M}odel for {S}cientific {T}ext}.
\newblock In \emph{Proceedings of the 2019 Conference on Empirical Methods in
  Natural Language Processing and the 9th International Joint Conference on
  Natural Language Processing (EMNLP-IJCNLP)}, pages 3615--3620, Hong Kong,
  China.

\bibitem[{Coley et~al.(2019{\natexlab{a}})Coley, Jin, Rogers, Jamison,
  Jaakkola, Green, Barzilay, and Jensen}]{coley2019graph}
Connor~W Coley, Wengong Jin, Luke Rogers, Timothy~F Jamison, Tommi~S Jaakkola,
  William~H Green, Regina Barzilay, and Klavs~F Jensen. 2019{\natexlab{a}}.
\newblock A graph-convolutional neural network model for the prediction of
  chemical reactivity.
\newblock \emph{Chemical science}, 10(2):370--377.

\bibitem[{Coley et~al.(2019{\natexlab{b}})Coley, Thomas~III, Lummiss, Jaworski,
  Breen, Schultz, Hart, Fishman, Rogers, Gao et~al.}]{coley2019robotic}
Connor~W Coley, Dale~A Thomas~III, Justin~AM Lummiss, Jonathan~N Jaworski,
  Christopher~P Breen, Victor Schultz, Travis Hart, Joshua~S Fishman, Luke
  Rogers, Hanyu Gao, et~al. 2019{\natexlab{b}}.
\newblock A robotic platform for flow synthesis of organic compounds informed
  by ai planning.
\newblock \emph{Science}, 365(6453):eaax1566.

\bibitem[{Collins et~al.(2020)Collins, Stout, Lim, Malerich, White, Madrid,
  Latendresse, Krieger, Szeto, Vu et~al.}]{collins2020fully}
Nathan Collins, David Stout, Jin-Ping Lim, Jeremiah~P Malerich, Jason~D White,
  Peter~B Madrid, Mario Latendresse, David Krieger, Judy Szeto, Vi-Anh Vu,
  et~al. 2020.
\newblock Fully automated chemical synthesis: toward the universal synthesizer.
\newblock \emph{Organic Process Research \& Development}, 24(10):2064--2077.

\bibitem[{Devlin et~al.(2019)Devlin, Chang, Lee, and
  Toutanova}]{devlin-etal-2019-bert}
Jacob Devlin, Ming-Wei Chang, Kenton Lee, and Kristina Toutanova. 2019.
\newblock \href {https://doi.org/10.18653/v1/N19-1423} {{BERT}: {P}re-training
  of {D}eep {B}idirectional {T}ransformers for {L}anguage {U}nderstanding}.
\newblock In \emph{Proceedings of the 2019 Conference of the North {A}merican
  Chapter of the Association for Computational Linguistics: Human Language
  Technologies}, pages 4171--4186, Minneapolis, Minnesota.

\bibitem[{Guo et~al.(2021)Guo, Ibanez-Lopez, Gao, Quach, Coley, Jensen, and
  Barzilay}]{guo2021correction}
Jiang Guo, A~Santiago Ibanez-Lopez, Hanyu Gao, Victor Quach, Connor~W Coley,
  Klavs~F Jensen, and Regina Barzilay. 2021.
\newblock Automated chemical reaction extraction from scientific literature.
\newblock \emph{Journal of chemical information and modeling}.

\bibitem[{Klucznik et~al.(2018)Klucznik, Mikulak-Klucznik, McCormack, Lima,
  Szymku{\'c}, Bhowmick, Molga, Zhou, Rickershauser, Gajewska
  et~al.}]{klucznik2018efficient}
Tomasz Klucznik, Barbara Mikulak-Klucznik, Michael~P McCormack, Heather Lima,
  Sara Szymku{\'c}, Manishabrata Bhowmick, Karol Molga, Yubai Zhou, Lindsey
  Rickershauser, Ewa~P Gajewska, et~al. 2018.
\newblock Efficient syntheses of diverse, medicinally relevant targets planned
  by computer and executed in the laboratory.
\newblock \emph{Chem}, 4(3):522--532.

\bibitem[{Kulkarni et~al.(2018)Kulkarni, Xu, Ritter, and
  Machiraju}]{kulkarni2018annotated}
Chaitanya Kulkarni, Wei Xu, Alan Ritter, and Raghu Machiraju. 2018.
\newblock An annotated corpus for machine reading of instructions in wet lab
  protocols.
\newblock In \emph{Proceedings of the 2018 Conference of the North American
  Chapter of the Association for Computational Linguistics: Human Language
  Technologies, Volume 2 (Short Papers)}, pages 97--106.

\bibitem[{Kuniyoshi et~al.(2020)Kuniyoshi, Makino, Ozawa, and
  Miwa}]{kuniyoshi2020annotating}
Fusataka Kuniyoshi, Kohei Makino, Jun Ozawa, and Makoto Miwa. 2020.
\newblock Annotating and extracting synthesis process of all-solid-state
  batteries from scientific literature.
\newblock In \emph{Proceedings of the 12th Language Resources and Evaluation
  Conference}, pages 1941--1950.

\bibitem[{Lowe(2017)}]{uspto_lowe}
Daniel Lowe. 2017.
\newblock \href
  {https://figshare.com/articles/dataset/Chemical_reactions_from_US_patents_1976-Sep2016_/5104873}
  {Chemical reactions from {US} patents (1976-2016)}.

\bibitem[{Lowe(2012)}]{Lowe2012ExtractionOC}
Daniel~M. Lowe. 2012.
\newblock {E}xtraction of {C}hemical {S}tructures and {R}eactions from the
  {L}iterature ({D}octoral {T}hesis).

\bibitem[{Mysore et~al.(2019)Mysore, Jensen, Kim, Huang, Chang, Strubell,
  Flanigan, McCallum, and Olivetti}]{mysore2019materials}
Sheshera Mysore, Zachary Jensen, Edward Kim, Kevin Huang, Haw-Shiuan Chang,
  Emma Strubell, Jeffrey Flanigan, Andrew McCallum, and Elsa Olivetti. 2019.
\newblock The materials science procedural text corpus: Annotating materials
  synthesis procedures with shallow semantic structures.
\newblock In \emph{Proceedings of the 13th Linguistic Annotation Workshop},
  pages 56--64.

\bibitem[{Nguyen et~al.(2020)Nguyen, Zhai, Yoshikawa, Fang, Druckenbrodt,
  Thorne, Hoessel, Akhondi, Cohn, Baldwin, and Verspoor}]{Nguyen2020ChEMUNE}
Dat~Quoc Nguyen, Zenan Zhai, Hiyori Yoshikawa, Biaoyan Fang, Christian
  Druckenbrodt, Camilo Thorne, Ralph Hoessel, S.~Akhondi, Trevor Cohn, Timothy
  Baldwin, and K.~Verspoor. 2020.
\newblock {ChEMU}: {N}amed {E}ntity {R}ecognition and {E}vent {E}xtraction of
  {C}hemical {R}eactions from {P}atents.
\newblock \emph{Advances in Information Retrieval}, pages 572 -- 579.

\bibitem[{Olivetti et~al.(2020)Olivetti, Cole, Kim, Kononova, Ceder, Han, and
  Hiszpanski}]{olivetti2020data}
Elsa~A Olivetti, Jacqueline~M Cole, Edward Kim, Olga Kononova, Gerbrand Ceder,
  Thomas Yong-Jin Han, and Anna~M Hiszpanski. 2020.
\newblock Data-driven materials research enabled by natural language processing
  and information extraction.
\newblock \emph{Applied Physics Reviews}, 7(4).

\bibitem[{O’Gorman et~al.(2021)O’Gorman, Jensen, Mysore, Huang, Mahbub,
  Olivetti, and McCallum}]{o2021ms}
Tim O’Gorman, Zach Jensen, Sheshera Mysore, Kevin Huang, Rubayyat Mahbub,
  Elsa Olivetti, and Andrew McCallum. 2021.
\newblock Ms-mentions: Consistently annotating entity mentions in materials
  science procedural text.
\newblock In \emph{Proceedings of the 2021 Conference on Empirical Methods in
  Natural Language Processing}, pages 1337--1352.

\bibitem[{Ravfogel et~al.(2021)Ravfogel, Taub-Tabib, and
  Goldberg}]{ravfogel2021neural}
Shauli Ravfogel, Hillel Taub-Tabib, and Yoav Goldberg. 2021.
\newblock Neural extractive search.
\newblock In \emph{Proceedings of the 59th Annual Meeting of the Association
  for Computational Linguistics and the 11th International Joint Conference on
  Natural Language Processing: System Demonstrations}, pages 210--217.

\bibitem[{Struble et~al.(2020)Struble, Alvarez, Brown, Chytil, Cisar,
  DesJarlais, Engkvist, Frank, Greve, Griffin et~al.}]{struble2020current}
Thomas~J Struble, Juan~C Alvarez, Scott~P Brown, Milan Chytil, Justin Cisar,
  Renee~L DesJarlais, Ola Engkvist, Scott~A Frank, Daniel~R Greve, Daniel~J
  Griffin, et~al. 2020.
\newblock Current and future roles of artificial intelligence in medicinal
  chemistry synthesis.
\newblock \emph{Journal of medicinal chemistry}, 63(16):8667--8682.

\bibitem[{Tamari et~al.(2021)Tamari, Bai, Ritter, and
  Stanovsky}]{tamari-etal-2021-process}
Ronen Tamari, Fan Bai, Alan Ritter, and Gabriel Stanovsky. 2021.
\newblock \href {https://doi.org/10.18653/v1/2021.eacl-main.187} {Process-level
  representation of scientific protocols with interactive annotation}.
\newblock In \emph{Proceedings of the 16th Conference of the European Chapter
  of the Association for Computational Linguistics: Main Volume}, pages
  2190--2202, Online. Association for Computational Linguistics.

\bibitem[{Tetko et~al.(2020)Tetko, Karpov, Van~Deursen, and
  Godin}]{tetko2020state}
Igor~V Tetko, Pavel Karpov, Ruud Van~Deursen, and Guillaume Godin. 2020.
\newblock State-of-the-art augmented nlp transformer models for direct and
  single-step retrosynthesis.
\newblock \emph{Nature communications}, 11(1):1--11.

\bibitem[{Valenzuela-Esc{\'a}rcega et~al.(2020)Valenzuela-Esc{\'a}rcega,
  Hahn-Powell, and Bell}]{valenzuela-escarcega-etal-2020-odinson}
Marco~A. Valenzuela-Esc{\'a}rcega, Gus Hahn-Powell, and Dane Bell. 2020.
\newblock \href {https://aclanthology.org/2020.lrec-1.267} {{O}dinson: A fast
  rule-based information extraction framework}.
\newblock In \emph{Proceedings of the 12th Language Resources and Evaluation
  Conference}, pages 2183--2191, Marseille, France. European Language Resources
  Association.

\bibitem[{Vaucher et~al.(2020)Vaucher, Zipoli, Geluykens, Nair, Schwaller, and
  Laino}]{vaucher2020automated}
Alain~C Vaucher, Federico Zipoli, Joppe Geluykens, Vishnu~H Nair, Philippe
  Schwaller, and Teodoro Laino. 2020.
\newblock Automated extraction of chemical synthesis actions from experimental
  procedures.
\newblock \emph{Nature communications}, 11(1):1--11.

\bibitem[{Zong et~al.(2020)Zong, Baheti, Xu, and Ritter}]{zong2020extracting}
Shi Zong, Ashutosh Baheti, Wei Xu, and Alan Ritter. 2020.
\newblock Extracting a knowledge base of covid-19 events from social media.
\newblock \emph{arXiv preprint arXiv:2006.02567}.

\end{thebibliography}
